# Image Inpainting Using AutoEncoder and Guided Selection of Predicted Pixels

Mohammad H. Givkashi[1]*, Mahshid Hadipour[2], Arezoo PariZanganeh[3], Zahra Nabizadeh[4], Nader Karimi[5], Shadrokh Samavi[6]

[1,2,3,4,5,6]Dept. of Elect. and Comp. Engineering, Isfahan University of Technology, Isfahan 84156-83111, Iran
[6]Dept. of Elect. and Comp. Engineering, McMaster University, L8S 4L8, Canada

*Abstract*— **Image inpainting is an effective method to enhance distorted digital images. Different inpainting methods use the information of neighboring pixels to predict the value of missing pixels. Recently deep neural networks have been used to learn structural and semantic details of images for inpainting purposes. In this paper, we propose a network for image inpainting. This network, similar to U-Net, extracts various features from images, leading to better results. We improve the final results by replacing the damaged pixels with the recovered pixels of the output images. Our experimental results show that this method produces high-quality results compare to the traditional methods.**

*Keywords*— *image inpainting, distortion mask, missing pixels, U-Net,*

## 1- Introduction

In recent years, the importance of improving the quality of digital images has been increased since they are highly applied in different fields of science. The image reconstruction includes substituting missing regions, removing undesired parts, or omitting either a text or sign from an image. One of the techniques which could be used to predict and replace the undesired pixels is image inpainting. There are two different approaches for image inpainting, including traditional [1-8] and learning-based methods [9-15].

### A. Traditional Methods

The traditional approaches are categorized as a diffusion-based and patch-based method where image reconstruction could be achieved by substituting noisy pixels using untouched ones. In the first method, destroyed pixels employ information of neighbors. On the other hand, substitution is applied by seeking potentially similar regions, and this is done by considering various criteria [1]. For example, the authors of [2] investigate the removal of line scratches from old films. Method of reference [2] detects image scratches using the information in frames and developed an approach in which missing pixels are predicted using P-Laplace. Also, the authors of [3] proposed an algorithm based on propagating an image smoothness estimator along the image gradient. They used the Fast Multipole Method (FMM) to disseminate image information. They estimated the lost pixels by using the weighted average in the neighborhood of the available pixels. Several methods based on Partial Differential Equations (PDE) have been introduced that are very time-consuming. A rapid convergence method with a combination of Richardson extrapolation and modified BSCB is proposed in [4].

In [5], Bertalmio et al. proposed a technique to combine image inpainting and texture synthesis that applied to the region of missing information. Patch-match is a new randomized algorithm for quickly finding approximate nearest neighbor matches between image patches that improve performance [6]. Another inpainting method highlights using a one-dimensional interpolation in four directions on various scales of an image. Then the result is accomplished by calculating the average of multi-scale paths [7]. In [8], a new procedure for image inpainting has contributed to the adaptive selection of neighbors of the missing pixel. The algorithm detects the location of the missing pixel on edges, removes invalid results, and uses the average value of the valid results. Their results show improvements in all metrics.

The above studies mostly employ methods with a heavy computational burden for image preprocessing or missing pixels prediction. However, these methods are suitable for handling images with small and thin lines. Moreover, a deep and high-level understanding of an image is not available. Hence, this motivated us to utilize deep neural networks or generative models.

---

\* The first three authors contributed equally to this work.

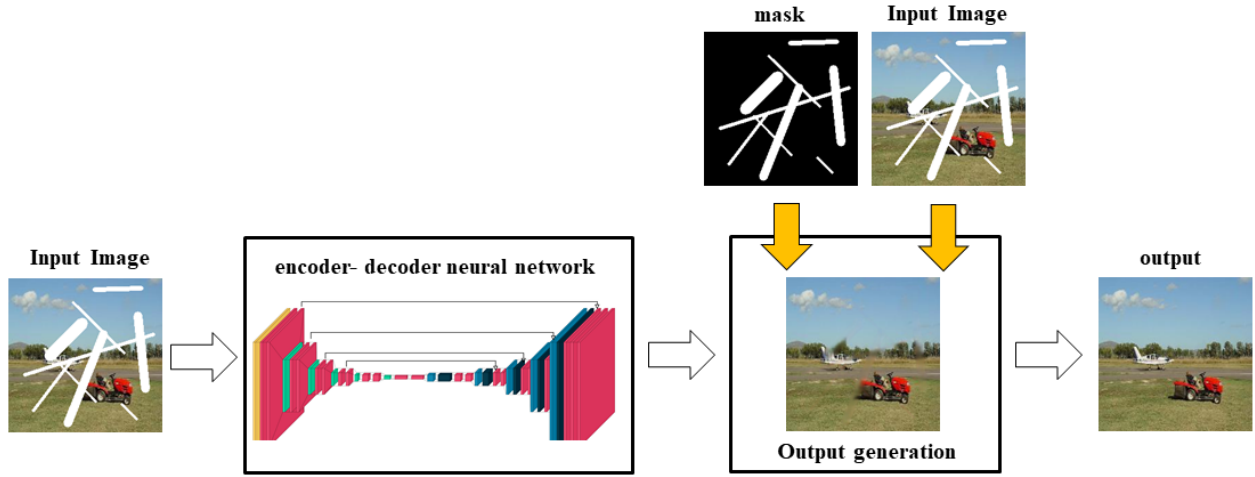

Figure. 1. Block diagram of the proposed method.

*B. Learning-based Methods*

For inpainting purposes, missing pixels have a mask. Hence, we refer to pixels as mask pixels and valid pixels. Sometimes, more information about neighboring pixels is required to recover the masked regions. In this case, learning-based methods such as deep neural networks (DNNs) and generative adversarial networks (GANs) may tackle this issue. As a result, some information about the image features is obtained, which may be used for image reconstruction. For instance, authors in [9] presented a method where context encoders are applied, leading to better results. Likewise, [10] suggests that using perceptual reconstruction losses at training time may improve the structure of networks like [9]. Most deep generative models are based on vanilla convolution, which uses valid and invalid pixels to extract features, causing visual artifacts such as color discrepancy and blurriness. According to the proposed method in [11], partial convolution can be utilized to provide networks with a better performance where features may only be achieved by considering valid pixels. Consequently, the propagated error from masked regions of an image is reduced.

In [12], filling the missing areas is divided into several phases to address the smaller parts. Next, it combines the results of all stages by using the LSTM structure. Nonetheless, this method seems inappropriate for irregular masks or those with different dimensions. Finally, in [13], the authors proposed a method, which applies both local and global features of images. This method can reconstruct images with arbitrary resolutions since it employs a convolutional neural network. Furthermore, to provide a more precise and compelling reconstruction of an image, it uses global and local context discriminators to discriminate an original picture from a recovered one.

The authors of [1] introduced utilizing a full-resolution residual block (FRRB) that updates the mask when required and fills missing regions much better than another algorithm. A strategy based on Multi-Scale Image Contextual Attention Learning (MUSICAL) is suggested by [14], where issues associated with poor information about the background are addressed. Also, downsampling layers are substituted by stride-1 dilated convolution layers, leading to better performance. Finally, a two-stage adversarial model Edge Connect is practiced [15]. An edge generator is applied to predict missing parts, and a network is developed by exploiting the knowledge of previous steps.

For image inpainting, there are different corruption masks. For example, some masks are block-based, and some are line-based, like what happens to old images. In this paper, we focus on repairing images that some of their pixels are removed. Since the width of the lines could be thick, it needs to have a global view of the image. For this purpose, a neural network is used to reconstruct the image by predicting the lost pixels. In the following, the proposed method is explained in section II, experimental results are shown in section III, and eventually, a concise conclusion is presented in section IV.

## 2- Proposed Method

Recently, applying neural networks has generated better results than traditional methods. The results have been improved for image inpainting by considering the whole image and its semantic relevance. But still, in these methods, the desired parameters have not improved enough. Therefore, we try to get better results by changing a basic network. This section will explain our proposed method and presents an overview of the inpainting network. The block diagram of our proposed method is shown in Fig. 1. As seen, the proposed method consists of two stages, Encoder- decoder neural network, Output generation.

*A. Network Design*
  *1) Encoder- decoder neural network*

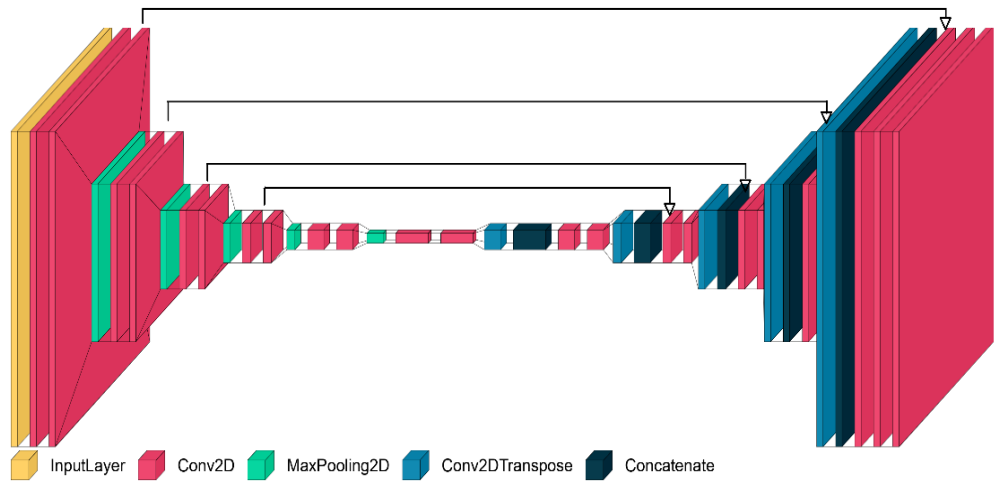

Figure. 2. Encoder-decoder structure of the proposed method.

Stage A in our proposed method includes an encoder-decoder network that is inspired by the U-Net network structure [16]. Compared to the network used in [17], the number of layers has been changed. This network is an extension and modification of the network presented by [16] in which upsampling operators have replaced pooling operators. High-resolution features from the concatenated path are combined with the upsampled output. A successive convolution layer can then learn to assemble a more precise output based on this information. One of the significant modifications of this model is the existence of feature channels in the decoder. These channels transfer the context information to the upsample layers. The architecture of the proposed neural is shown in Fig. 2.

We increased the number of layers due to the size of the input images. The base network uses fewer input layers than the network that is implemented here.

*2) Output generation*

Due to the structure of learning networks, all the pixels of the images are predicted. However, since the quality of the neural network-generated image does not equal the actual picture, all destroyed and non-distorted pixels are changed. Hence, no matter how small and imperceptible these changes are, they will affect the final results. To solve this problem and reduce the final error, we only replace the damaged pixels in the input image with the reconstructed pixels in the output image of the network. For this purpose, the missing pixels are detected from the input mask. Then the detected pixels are extracted from the model output image and used in the input image.

## 3- Experiments

*A. Training process*

We use at least 20000 images of the ADE20K [18] dataset to train our model. To evaluate our model, we use the test images of this dataset and the KODAK [19] dataset. The size of the training dataset images is changed to 224×224×3. Also, augmentation methods have been used to increase the number of training data. First, the images are randomly selected to be flipped vertically or horizontally for augmenting the dataset. Then a random set of images is selected for rotation. The rotation is 90, 180, or 270 degrees. We used Adam optimizer [20], a batch size of 64, and 100 epochs of training.

*B. Implementation*

Our algorithm is implemented by extending the existing Keras library [21], inspired by [22]. The entire network inference on a 224×224×3 image runs on a single Nvidia GeForce RTX 3090 GPU. Twenty-four images from the KODAK data set and the same number of ADE20K are used to report the final results, and the numbers reported in the results are the average values obtained from each image. As mentioned earlier, to generate the destroyed images, pixels of the image are destroyed by random masks. The thickness of these masks is a parameter that can affect the final results.

TABLE 1  Comparison of our method and [7] with fix masks.

| Input image | Mask | Method [7] | Our method |
|---|---|---|---|
|  |  |  |  |
| PSNR(dB) | - | 38.7724 | 38.06394 |
| SSIM | - | 0.9793 | 0.9970 |
|  |  |  |  |
| PSNR(dB) | - | 34.6655 | 34.08846 |
| SSIM | - | 0.9829 | 0.9942 |
|  |  |  |  |
| PSNR(dB) | - | 33.2983 | 37.79142 |
| SSIM | - | 0.9797 | 0.9924 |
|  |  |  |  |
| PSNR(dB) | - | 35.4859 | 35.40415 |
| SSIM | - | 0.9829 | 0.991 |

TABLE 2  Comparison of our method and [7] with variable and narrow masks.

| Input image | Mask | Method [7] | Our method |
|---|---|---|---|
| 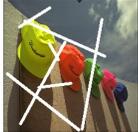 | 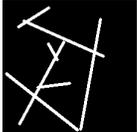 | 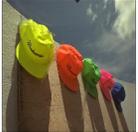 | 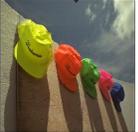 |
| PSNR(dB) | - | 33.9898 | 33.9441 |
| SSIM | - | 0.944238 | 0.9900 |
| 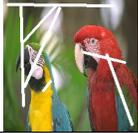 | 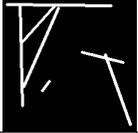 | 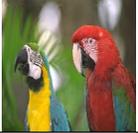 | 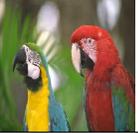 |
| PSNR(dB) | - | 33.3053 | 33.2863 |
| SSIM | - | 0.950598 | 0.9872 |
| 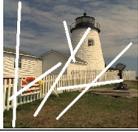 | 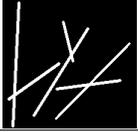 | 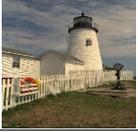 | 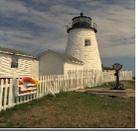 |
| PSNR(dB) | - | 28.9407 | 28.4414 |
| SSIM | - | 0.949912 | 0.9757 |
| 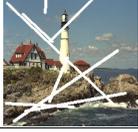 | 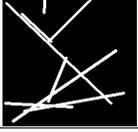 | 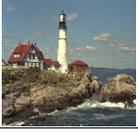 | 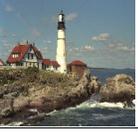 |
| PSNR(dB) | - | 29.9063 | 30.4002 |
| SSIM | - | 0.93518 | 0.9700 |

TABLE 3  Comparison of our method and [7] with variable and thick masks.

| Input image | Mask | Method [7] | Our method |
|---|---|---|---|
| 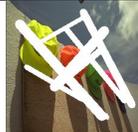 | 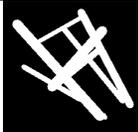 | 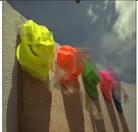 | 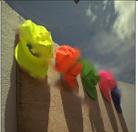 |
| PSNR(dB) | - | 25.5943 | 26.6392 |
| SSIM | - | 0.7932 | 0.9550 |
| 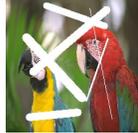 | 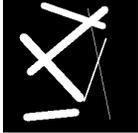 | 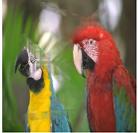 | 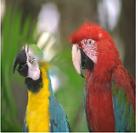 |
| PSNR(dB) | - | 26.1108 | 26.6400 |
| SSIM | - | 0.8838 | 0.9677 |
| 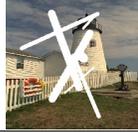 | 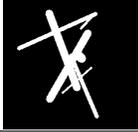 | 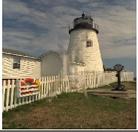 | 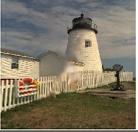 |
| PSNR(dB) | - | 26.7169 | 28.3099 |
| SSIM | - | 0.9070 | 0.9616 |
| 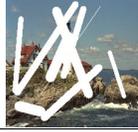 | 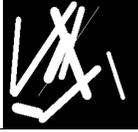 | 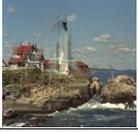 | 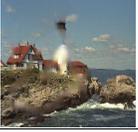 |
| PSNR(dB) | - | 23.1087 | 23.9993 |
| SSIM | - | 0.8088 | 0.9084 |

As mentioned earlier, the resulting masks are applied to the pictures, which accidentally damage part of the image. The thickness of this mask is a parameter that affects the final results.

*C. Mask dataset*

We produce masks the same size as the input image using a code similar to [22], which can change the size and dimensions of the number of pixels destroyed. We used it in several ways to observe the effect of mask dimensions on the results obtained in different methods.

*D. Comparison*

Several experiments are performed, which we report in this section to evaluate the proposed structure comprehensively.

We compared the result of our work with an interpolation method [7]. Evaluation of these methods has been performed by using PSNR and SSIM.

In the first experiment, we used narrow masks to destroy pixels in the center of images. In these masks, the width of all lines is the same. In this case, the network was not well trained, and the results of the interpolation method, which uses neighbor pixels of the destroyed regions for reconstruction, are better or slightly different from the output of our approach. The results are reported in Table 1.

In the next step, we consider the width of lines in the masks to be variable and changes from 1 to 8 pixels. We also do not set any restrictions on the location of the masks. The results of this test are shown in Table 2. In this case, the results of the two methods are almost similar.

To more accurately observe the effect of the masks on the results obtained from the two methods, we produced masks with different widths and thicker than the previous step. We see that the interpolation method cannot completely restore the lost pixels in this method .The results are reported in Table 3.

To better compare, the average results of these three situations are reported in Table 4. We see that in cases where the thickness of the produced masks is not so great, in the limited number of images, the PSNR of the [7] method was better than our method.
However, the interpolation method cannot have a good answer in the other case. But since the neural network has a local and global view, the SSIM of our proposed method is better than the method in [7] in all situations.

TABLE 4  Comparison of our method and [7] in different cases.

|  |  | ADE20K | | KODAK | |
|---|---|---|---|---|---|
|  |  | PSNR(dB) | SSIM | PSNR(dB) | SSIM |
| Method [7] | 1 | 35.9096 | 0.9841 | 36.2209 | 0.9811 |
|  | 2 | 31.6473 | 0.95 | 32.5392 | 0.9483 |
|  | 3 | 26.8449 | 0.8981 | 27.3865 | 0.8762 |
| Our method | 1 | 35.3546 | 0.9908 | 36.1792 | 0.9916 |
|  | 2 | 31.6473 | 0.9781 | 32.5392 | 0.9779 |
|  | 3 | 27.9121 | 0.95265 | 28.5956 | 0.9495 |

## 4- Conclusion

Inpainting is a widely used image processing application to improve image quality or generate new images by removing objects from the scene. This paper showed that both interpolation and learning methods could effectively perform inpainting. However, their final results depend on the structure and dimensions of the distorted regions. Our experiments showed that learning networks are more capable of adapting to situations. For example, it can more accurately recover damaged areas caused by thicker scratches.

Furthermore, since the global and local information is used in our method, the reconstructed images could have preserved the structure of pictures. The calculated SSIM also approves this claim.

We intend to apply the proposed method to video sequences for future work. We can also use this platform to denoise medical images [23-25].